# Who supervises the supervisor? Data drift detection in production using deep feature embeddings with applications to workpiece inspection

Michael Banf, Gregor Steinhagen

*Fabforce GmbH & Co. KG, 57250 Nepthen, Germany*

ABSTRACT

The automation of condition monitoring and workpiece inspection plays an essential role in maintaining high quality as well as high throughput of the manufacturing process. To this end, the recent rise of developments in machine learning has lead to vast improvements in the area of autonomous process supervision. However, the more complex and powerful these models become, the less transparent and explainable they generally are as well. One of the main challenges is the monitoring of live deployments of these machine learning systems and raising alerts when encountering events that might impact model performance. In particular, supervised classifiers are typically build under the assumption of stationarity in the underlying data distribution. For example, a visual inspection system trained on a set of material surface defects generally does not adapt or even recognize gradual changes in the data distribution - an issue known as "data drift" - such as the emergence of new types of surface defects. This, in turn, may lead to detrimental mispredictions, e.g. samples from new defect classes being classified as non-defective. To this end, it is desirable to provide real-time tracking of a classifier's performance to inform about the putative onset of additional error classes and the necessity for manual intervention with respect to classifier re-training. Here, we propose an unsupervised framework that acts on top of a supervised classification system, thereby harnessing its internal deep feature representations as a proxy to track changes in the data distribution during deployment and, hence, to anticipate classifier performance degradation.

## 1. Introduction

Today's increased level of automation in manufacturing also requires the automation of material quality inspection with as little as possible human intervention. To stay competitive while meeting industry standards, companies strive to achieve both quantity and quality in production without compromising one over the other. However, manual quality inspection or extensive equipment tests typically allow only for the analysis of individual samples from a given batch of products. With the emergence of vast improvements in the area of Artificial Intelligence, companies begin to employ such technologies during the production cycle to automate quality inspection, as well as monitor machine conditions, thereby minimizing human intervention, and optimizing factory capacities. Accordingly, a plethora of machine learning based condition monitoring and workpiece inspection methodologies and applications have been proposed, including gearbox Jing et al., 2017 or rotary machinery analyses Chen et al., 2015, as well as bearing ke Wen et al., 2018; Lu and ling Chen, 2021 or steel surface defect detection Deshpande et al., 2020; Lv et al., 2020; Huang et al., 2018; Tao et al., 2018; Konovalenko et al., 2020; Ferguson et al., 2018, just to name a few. For a more in-depth survey on machine learning for condition monitoring we refer the reader to Yang et al., 2020; Fang et al., 2020 or Qi et al., 2020.

In contrast, model performance tracking of such classification systems during production is an area less explored.

However, as discussed in Klaise et al., 2020, the life cycle of a machine learning system does extend beyond its deployment, and one of the big challenges is to design systems that are able to monitor live deployments and take appropriate actions on encountering model performance impacting events Diethe et al., 2019.

One such event may be a gradual or abrupt drift in the data distribution. In the case of a deployed machine learning model, this would be the change between real-time production and the baseline data set used for initial model training Žliobaitė et al., 2016. In general, supervised classification systems are trained under the assumption of stationarity in the underlying - in most cases - latent, i.e. non directly observable, data distribution. For example, a model that has been trained as a visual inspection classifier for a set of putative material surface defect types, generally will not adapt or even recognize gradual changes in the data distribution, such as the emergence of a new type of surface defect, which, as a consequence, may then be mispredicted as non-defective. These are critical weaknesses that must be accounted for in deployed classifiers in order to avoid model performance degradation without the user even noticing.

It has been proposed that, in the absence of labels for live data it is critical to monitor the statistics of input data and output predictions as these can serve as a proxy for model performance Breck et al., 2017. A monitoring system requires functionality to determine when significant changes to data and predictive distributions happen, also known as drift detection Dries and Rückert, 2009; Bach and Maloof, 2010; Žliobaitė et al., 2016; Gama et al., 2014; Chen et al., 2015; Minku et al., 2010; Webb et al., 2015; Lipton et al., 2018; Rabanser et al., 2019. A related task is to identify

---

m.banf@fabforce.com (M. Banf); g.steinhagen@fabforce.com (G. Steinhagen)
ORCID(s):

incoming data points which fall outside the training data distribution, referred to as outlier detection. These questions are statistical in nature and often require separate models which makes it more difficult to provide general solutions Klaise et al., 2020. Successful drift detection may be used to inform a user that the ongoing production data should be inspected and a deployed model may need to be retrained on an augmented training dataset.

Here, we propose a conceptional approach to address the problem of monitoring model performance during production by autonomously detecting data drift in an unsupervised fashion (see figures 1 and 2). In particular, we want our approach to identify drift in the data that may cause false negative predictions or type II errors, that is newly introduced defect types being classified as "non-defect" by the original classifier. Hence, our framework harnesses the classifier's learned internal feature representations in order to:

- track and identify anomalies within the latent representation space as the most natural way to track the data sample space without imposing an implicit probability distribution

- save memory and training time by an implicit dimensionality reduction of the input feature space

- allow for the usage of efficient, edge computing capable outlier detection algorithms for drift recognition.

We perform a series of experiments on two datasets to model different types of data drift as well as drift severity in a typical defect classification scenario, i.e. automated, visual quality inspection.

## 2. Methods

### 2.1. Definition of a latent representation space using deep feature embeddings

The main rationale of our proposed framework is to define a latent representation space by extracting - per training sample - the deep feature embeddings provided by the original supervised classifier, and, subsequently, comparing the feature representation of any new production sample with the set of training representations to identify putative outliers that might indicate a shift in the underlying data distribution, e.g. the emergence of a new defect type for which the original classifier has not been trained and may not correctly classify. This way, we may be able to track changes within the representation space, although - given the discriminative nature of classifier systems - we may not a priori assume that the learned feature representation space constitutes any type of continuous probability distribution.

We extract the sample based feature representation of the penultimate layer, assuming the classifier to be a typical feed forward type deep neural network (see figure 3). Here, we demonstrate our approach for a classification model trained to distinguish between two classes, i.e. "defect" and "non-defect". However, the same rationale can be applied and extended to arbitrary multi-class setups.

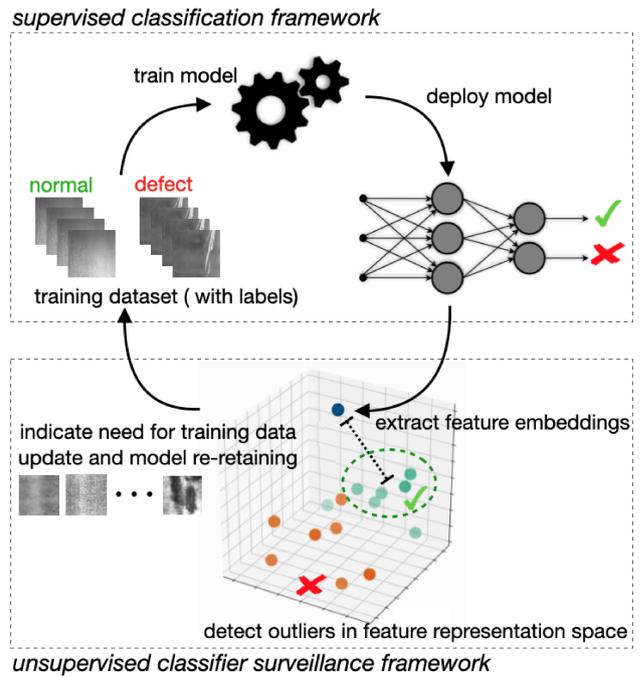

**Figure 1**: The machine learning system's lifecycle completed. Our framework monitors model performance via latent space representation tracking during production and informs about the need for training dataset augmentation and re-training.

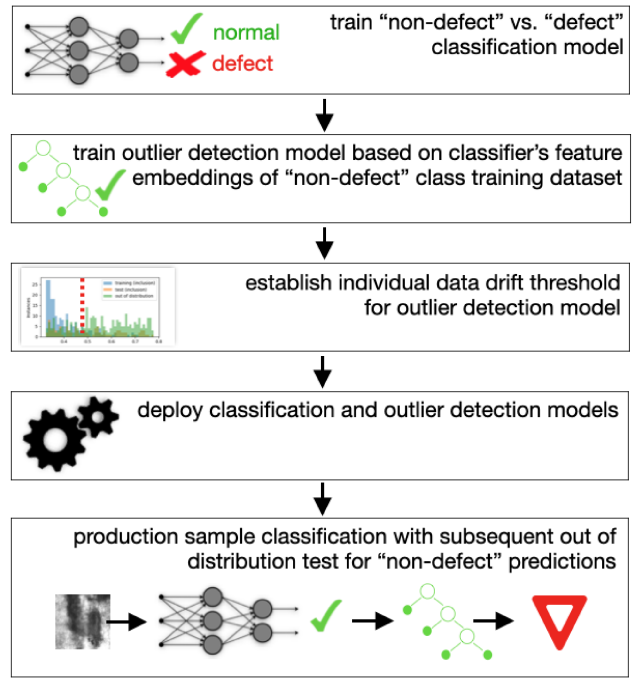

**Figure 2**: Overview of our proposed framework's workflow from classifier and subsequent out of distribution model training to deployment and usage in production.

### 2.2. Measuring data drift using Isolation Forests

In principle, given our latent space representation, a variety of anomaly detection algorithms may be used to detect outliers in latent space, ranging from statistical based methods to more recent developments in the area of Deep Learning, such as (Variational) Autoencoders Thudumu et al., 2020. Here, we chose Isolation Forests Liu et al., 2008, 2012; Hariri et al., 2021, an unsupervised, scalable and non-parametric outlier detection algorithm with comparatively low computational and memory requirements, making it ideal for on-edge computing applications Domingues et al., 2018. In addition, one of the major steps in recent outlier detection algorithms such as autoencoders is the implicit creation of a reduced feature representation space Rabanser et al., 2019, a putative computationally and training data expensive procedure that we avoid by directly applying Isolation Forests on the feature representations already learned and provided by the classifier as discussed above.

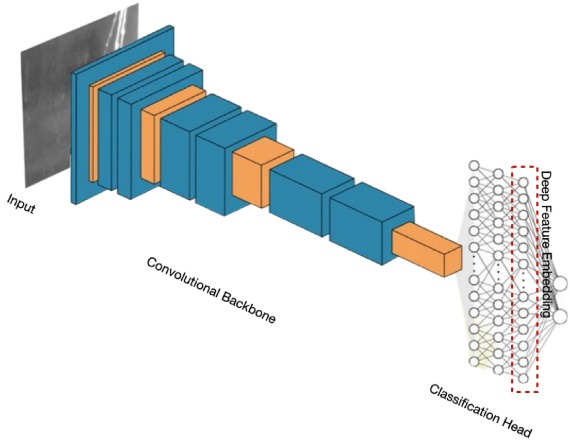

Figure 3: Example of a typical feed forward convolutional neural network with several interspersed convolutional (blue) and pooling (orange) layers, followed by multiple dense layers and a two class softmax layer to close. Extraction of the deep feature embedding from the classification head's penultimate layer (Image adapted from Hoeser and Kuenzer, 2020).

The main rationale of the Isolation Forest framework is to randomly generate binary tree structures from subsets of the training data and identify anomalies in each subset as instances with the shortest average path length in a given tree, i.e. instances that can be separated from the remaining training set with the least number of splits (see figure 4 for an illustration).

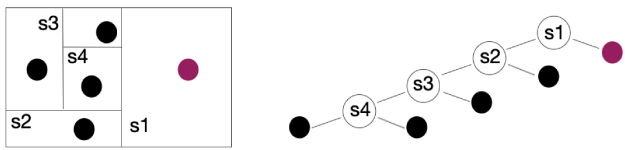

Figure 4: Toy example of identifying an anomalous point (orange) in 2D space by the number of tree splits needed to isolate it from the population (blue).

A single tree is created by: i) sampling a subset from the training data; ii) randomly choosing a splitting feature per node; and iii) randomly selecting a splitting value from a uniform distribution, spanning from the minimum to the maximum value of the feature selected in the step ii). Steps ii) and iii) are then repeated recursively, in theory, until all instances from the subset are "isolated" in individual leaf nodes. In practice, however, a height limit is generally applied, based on the assumption that outliers may be easier to isolate in leaf nodes and thus, on average, do require fewer random splits, resulting in a shorter path length from the root node to the leaf node. If, on building the tree, a height limit is applied, some leaf nodes will end up with more training instances than others. Therefore, an ensemble of tree models is trained, with outlier scores being averaged across their individual outputs to reduce the variance of the model. The outlier score $S$ for a particular instance $x$ is then computed as a function of the average path length from the root to the leaf node compared to the total number $m$ of training instances across all constructed trees, i.e.:

$$S(x, m) = 2^{\frac{-E(h(x))}{c(m)}},$$

with $E(h(x))$ denoting the average search height for $x$ across all trees, and $c(m)$ being a normalization constant for the training data subset of size $m$, defined as the average depth in an unsuccessful search in a Binary Search Tree:

$$c(m) = 2H(m-1) - 2(m-1)/m$$

Here, $H$ is referred to as the harmonic number, which can be estimated by $H(i) = ln(i) + \gamma$, with $\gamma$ denoting the Euler's constant Liu et al., 2012. As a consequence, if $E(h(x)) << c(m)$, then $S(x, m) = 1$, that is $x$ will most certainly be an outlier compared to the remaining $m$ instance. In contrast, if $E(h(x)) \approx c(m)$, then $S(x, m) \approx 0.5$. In an unsupervised setting, the number of hyper-parameters for the Isolation Forest model reduces to selecting the number of trees as well as the training subset sampling size.

Since we are primarily interested in the detection of type II errors, that is newly introduced defect types being classified as "non-defect" by the original classifier, we use the latent space feature representations of the "non-defect" training dataset to train an Isolation Forest as data drift detection model for "non-defect" predictions. Hence, during deployment, if a production sample has been classified as "non-defective", our framework performs an outlier detection based on the sample's feature representation using the "non-defect" Isolation Forest model. Depending on an estimated threshold, the distance of the sample's outlier score with respect to the training data may then be used as an indicator of the validity of the classifier's prediction. Predictions, so identified as potentially invalid, may then be utilized as indicators of conceptional inability of the classifier to recognize novel types of defects, and, hence, potential drift in the data (see figure 5).

### 2.3. Median absolute deviation data drift thresholding and classifier re-training

Given a trained data drift detector, we need to define an individual threshold on when to flag a production sample as a putative outlier for manual inspection. Here we do not impose a prior probability on the class-specific distribution of outlier scores from the training set, but propose a common heuristic, that is to calculate the median absolute deviation (MAD) Leys et al., 2013; Rousseeuw and Croux, 1993. In brief, for a univariate data set $X_1, X_2, ..., X_n$, the MAD is defined as $median(\|X_i - \overline{X}\|)$, i.e. the median of the absolute deviations from the data's median $\overline{X}$. Hence, the MAD is a robust measure of statistical dispersion and commonly used thresholds for outlier and, hence, drift identification in a distribution are 3.5 MADs.

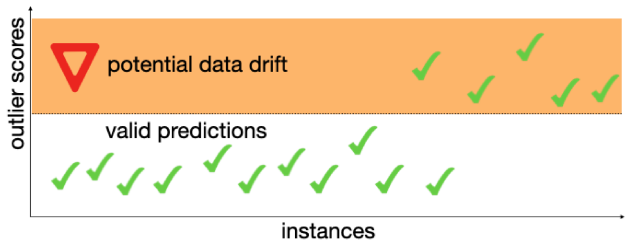

**Figure 5**: Outlier scores may be used to flag a classifier's "non-defect" predictions as potentially invalid and to identify data drift.

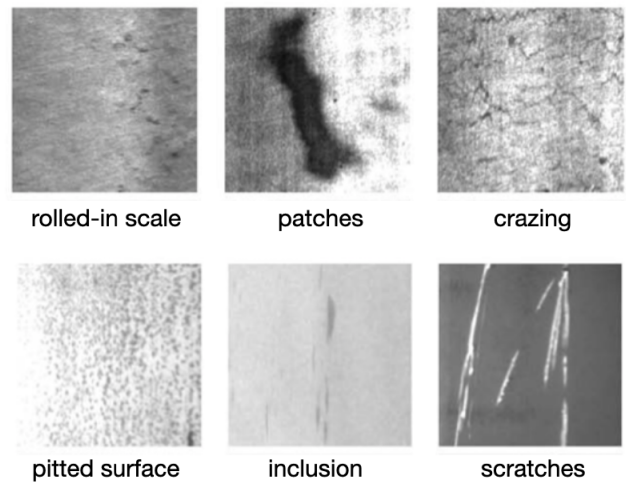

**Figure 6**: Overview of the six kinds of typical surface defects of the hot-rolled steel strip in the NEU benchmark dataset Lv et al., 2020.

Each so identified outlier may be flagged for further inspection. Given that a growing number of instances - so classified to constitute outliers - would be observed, these instances should be manually inspected to examine the emergence of data drift. If confirmed, all samples containing a new type of error class should be combined with the original training data in order to retrain the classifier, either to augment the "defect" class in the original binary classification model, or to define additional defect classes and change the classification model into a multi-class setup.

## 3. Results and Discussion

We evaluated our approach for two typical in production quality inspection tasks, i.e. visual inspection and classification of steel surface defects.

### 3.1. Visual inspection of steel surface defects on the NEU benchmark dataset

As a first experiment, we trained a classifier for a visual inspection task of steel surface defects. Hence, we resorted to a typically used benchmark dataset, i.e. the surface defect database from the North-Eastern University (NEU), China Lv et al., 2020. The dataset consisted of six kinds of typical surface defects of the hot-rolled steel strip, i.e., rolled-in scale, patches, crazing, pitted surface, inclusion and scratches (see figure 6), including in total 1800 grayscale images of 200 x 200 dimensionality with 300 samples per class. We separated the set of samples per class into 200 images for training and 100 for testing, respectively.

As classifier, we used a feed forward convolutional neural network based on the "MobileNet" architecture Sandler et al., 2018, given it's suitability as a go-to model for app and edge deployment Ahmed and Bons, 2020. Further, given the relative small size of our training data per class, i.e. 200 samples each, we applied "transfer learning" Abu et al., 2021, that is, the original model had been pre-trained on the large ImageNet database Russakovsky et al., 2015, and, subsequently, we customized and fine-tuned the model for our prediction task by freezing the weights of the convolutional backbone, and replacing the classification head with a customized one (see figure 3 for a general illustration) with a two class output, i.e. "non-defect" vs "defect". We choose the penultimate layer - which later served as the latent representation layer for feature extraction (see figure 3) - to be a dense layer containing 1024 neurons. Model training was performed with a validation split of 20 percent on the training dataset.

Given that the original benchmark dataset only consisted of six defect classes, missing a general "non-defect" class, we needed to define such a non-defect class for our experiments. In order to evaluate the robustness of our approach, that is to investigate the putative dependencies of performance on the specific distributions of selected class pairs, we used the six defect classes to create five representative class pairs labelled as "non-defect" and "defect" classes (see 1) and performed individual experiments on each pair. The test sets, i.e. 100 images each, of the respective remaining four classes then served as additional defect types, denoted as out-of-distribution (OOD) data, to be introduced during our simulated production experiments. Table 1 shows the results of our five experiments. Note that the performances of the original binary classifiers, each trained for a given pair of classes, were measured to be $\geq$ 99 percent on the training and validation datasets across experiments.

Since we were particularly interested in the proportion of type II errors, that is newly introduced defect types being classified as "non-defect" by the classifier, we classified instances from the test sets of the four remaining classes. Throughout the five experiments, we observed type II error rates ranging between 36 and 82 percent (see table 1, column 2). This confirmed the necessity for a classifier supervision framework to at least inform a user about such critical events of covert non-defect mispredictions.

**Table 1**
Experiment specific class pair permutations and Kolmogorov-Smirnov tests between "non-defect" outlier score distributions

| Training classes<br>Non-defect : Defect | type II errors<br>(false neg.)<br>on OOD | Outlier score<br>dist. similarity<br>Training<br>vs test | Training<br>vs OOD | Outlier score<br>threshold | Detection rate<br>of type II errors |
| --- | --- | --- | --- | --- | --- |
| Inclusion : Scratches | 58 % | $p< 2.5e^{-12}$ | $p< 1.9e^{-124}$ | 0.46 | 99 % |
| Patches : Crazing | 62 % | $p< 1.5e^{-5}$ | $p< 5.5e^{-16}$ | 0.48 | 77 % |
| Inclusion : Crazing | 36 % | $p< 4.1e^{-8}$ | $p< 4.4e^{-16}$ | 0.44 | 99% |
| Inclusion : Patches | 62 % | $p< 2.5e^{-8}$ | $p= 0$ | 0.46 | 100 % |
| Rolled-in scale : Crazing | 82 % | $p< 2.6e^{-12}$ | $p< 3.5e^{-260}$ | 0.47 | 100 % |

Subsequently, we learned our Isolation Forest based drift detection framework using the per sample latent representations extracted from the classifier's penultimate layer, thereby setting the number of trees to be 100. Using a non-parametric and distribution-free two-sample Kolomogorov-Smirnov test Peacock, 1983, we then examined the differences in the resulting distributions of outlier scores between the training and test datasets for the respective "non-defect" classes, as well as the OOD based distributions (see table 1; Note that a smaller p-value is a higher indicator that the Null-hypothesis, i.e. the two samples have been produced by the same distribution, may be rejected). Table 1 highlights that in all five experiments, the "non-defect" based outlier distributions of training and test data were more similar - with respect to shape and location - than the OOD distribution of "non-defect" predictions. As mentioned before, a data drift threshold was defined as 3.5 MADs from the median with respect to the "non-defect" training data based outlier score distribution, allowing for detection rate of invalid "non-defect" predictions (i.e. type II errors) of on average 95 percent (see table 1, column 6) and, thereby, the identification of data drift (see figure 7, bottom panel).

Finally, we assumed that different penultimate layer sizes may have separate effects on the classifier and the supervision model performances. Hence, we further performed a series of qualitative experiments in which the classifier as well as the data drift detectors were trained with varying latent representation sizes, i.e. different penultimate layer sizes, ranging from dimensionalities of 8 to 1024 neurons. We observed that in all five experiments, while classifier performance remained relatively stable, the data drift detection on the introduced OOD datasets typically performed poorly for very small penultimate layers (see figure 7, panel a). Hence, for practical deployment of a supervision framework, considerations on how to dimension the size of the feature representation space have to be taken into account.

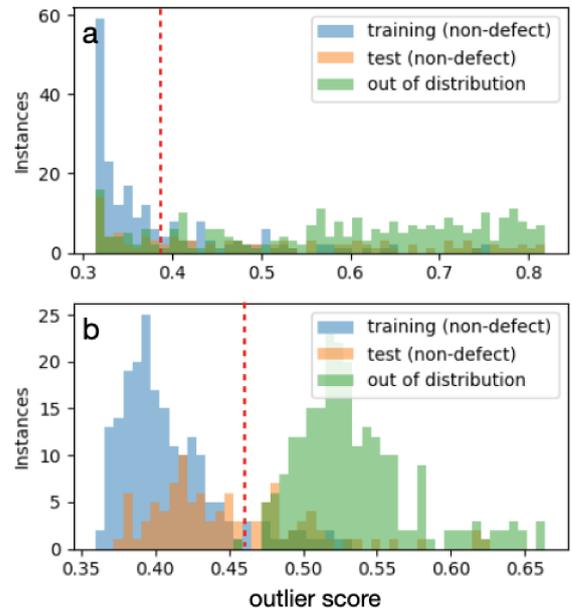

**Figure 7**: Outlier scores based on isolation forest trained on samples of inclusion ("non-defect") using per sample latent representation of classifier trained on inclusion ("non-defect") vs. scratches ("defect"). Latent representation spaces (penultimate layer) sizes were 8 (a) and 1024 neurons (b). Dashed red vertical lines highlight data drift thresholds, defined as 3.5 MADs from the median with respect to the "non-defect" training data based outlier score distribution. Compared to the 8 neuron model, the drift detector based on 1024 neurons allows for a clear separation between valid and invalid "non-defect" predictions and, thereby, the identification of data drift.

### 3.2. Visual inspection of steel surface defects on the Xsteel benchmark dataset

As a second experiment, we resorted to another, very recent, benchmark dataset for steel surface defects, the Xsteel surface defect dataset Feng et al., 2021. The dataset also consisted of typical surface defects of the hot-rolled steel strip, including in total 1360 grayscale images of 128 x 128 dimensionality, including 238 (slag) inclusions, 397 red iron sheet, 122 iron sheet ash, 134 (surface) scratches, 63 oxide scale of plate system, 203 finishing roll printing and 203 oxide scale of temperature system (see figure 8). We separate the set of samples per class into approximately 80 percent of the images for training and 20 percent for testing, respectively.

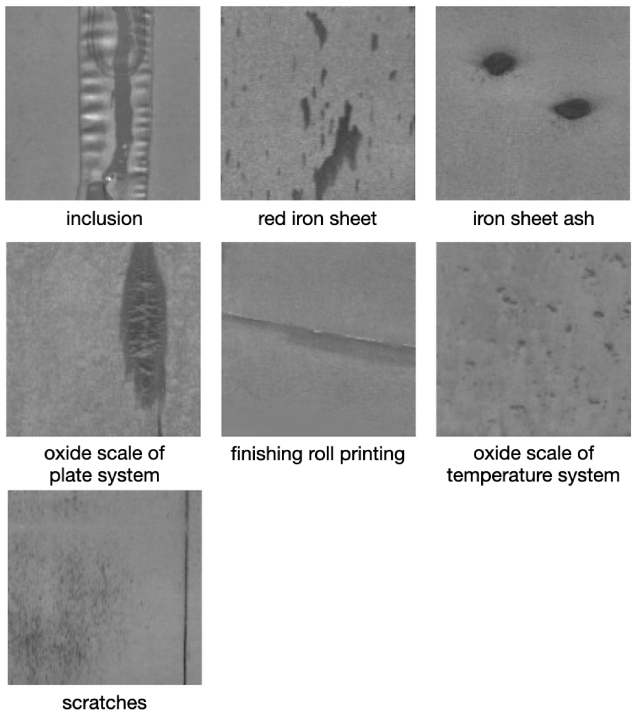

**Figure 8**: Overview of the seven kinds of surface defects of the hot-rolled steel strip in the XSteel benchmark dataset Feng et al., 2021.

As classifier, we used the same, pre-trained, feed forward convolutional neural network based on the "MobileNet" architecture Sandler et al., 2018, as in our first experiment, except that we reduced the input dimensions from 200 by 200 to 128 by 128. Model training was performed with a validation split of 20 percent on the training dataset. As in our first experiment, in order to evaluate the robustness of our approach, that is to investigate the putative dependencies of performance on the specific distributions of selected class pairs, we used the seven defect classes to create five representative class pairs labelled as "non-defect" and "defect" classes (see 1) and performed individual experiments on each pair. The test sets of the respective remaining five classes then served as additional defect types, denoted as out-of-distribution (OOD) data, to be introduced during our simulated production experiments. Table 1 shows the results of our five experiments. Note that the performances of the original binary classifiers, each trained for a given pair of classes, were measured to be ≥ 98 percent on the training and validation datasets across experiments.

Throughout all experiments, type II error rates were ranging between 31 and 65 percent (see table 2, column 2) and, as shown in table 2 and figure 9, the "non-defect" based outlier distributions of training and test data were, again, found to be more similar than the OOD distribution of "non-defect" predictions. As before, a data drift threshold for the Isolation Forest based drift detection framework was defined as 3.5 MADs from the median with respect to the

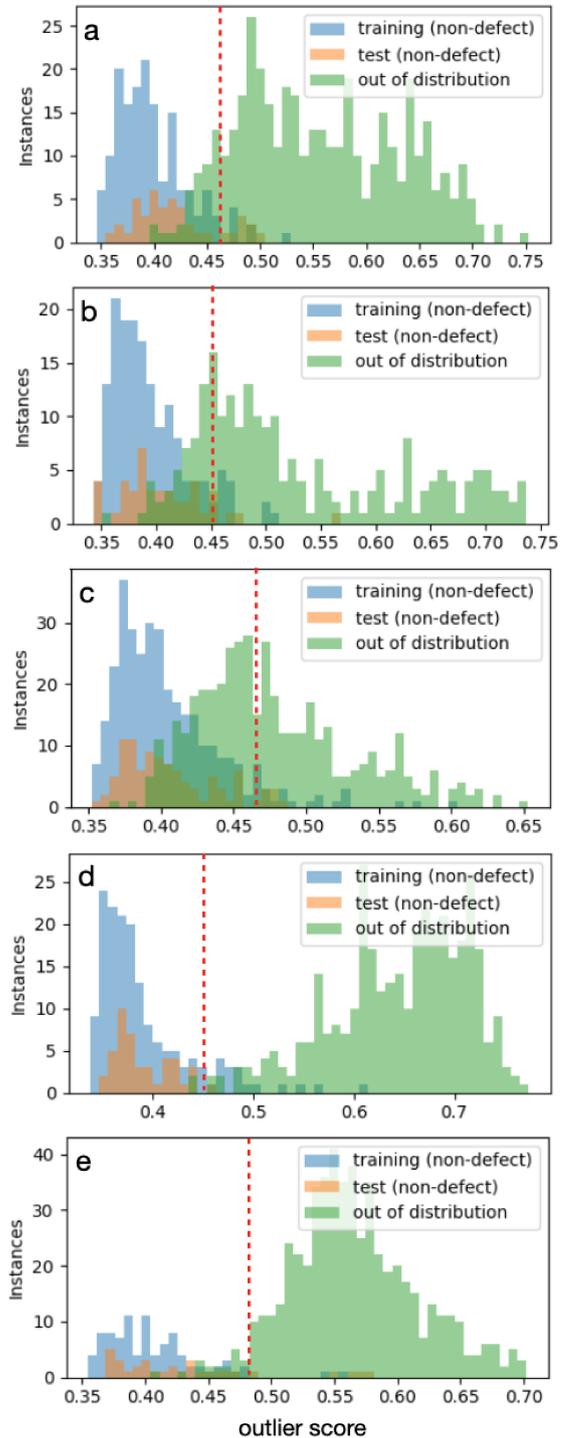

**Figure 9**: Outlier scores based on isolation forest trained on samples of the respective "non-defect" classes ( a) and b) finishing roll printing, c) red iron sheet, d) oxide scale of temp. system, e) iron sheet ash) using per sample latent representation of the respective trained classifier. Dashed red vertical line highlights data drift threshold, defined as 3.5 MADs from the median with respect to the "non-defect" training data based outlier score distribution.

"non-defect" training data based outlier score distribution,

Table 2
Experiment specific class pair permutations and Kolmogorov-Smirnov tests between "non-defect" outlier score distributions

| Training classes<br>Non-defect : Defect | type II errors<br>(false neg.)<br>on OOD | Outlier score dist. similarity | | Outlier score threshold | Detection rate of type II errors |
|---|---|---|---|---|---|
| | | Training vs test | Training vs OOD | | |
| Finished roll printing : Iron sheet ash | 48 % | $p< 0.0033$ | $p= 4.4e^{-102}$ | 0.47 | 89 % |
| Finished roll printing : Inclusion | 31 % | $p< 0.04$ | $p< 1.1e^{-61}$ | 0.45 | 76 % |
| Red iron sheet : Scratches | 65 % | $p< 0.81$ | $p= 0$ | 0.47 | 49 % |
| Oxide scale of temp. system : Oxide scale of plate system | 41 % | $p< 0.07$ | $p< 2.9e^{-137}$ | 0.44 | 99 % |
| Iron sheet ash : Oxide scale of plate system | 58 % | $p< 0.031$ | $p< 3.3e^{-16}$ | 0.48 | 97 % |

resulting in a detection rate of false negative predictions of on average 82 percent (see table 2, column 6).

## 4. Conclusion

One of the main challenges in the automation of condition monitoring and workpiece inspection for high quality, high throughput manufacturing is the monitoring of live deployments of assistive machine learning systems to track model performance.

Here, we addressed this less explored field of production model performance monitoring and proposed an unsupervised framework that acts on top of an existing, supervised classification system, thereby harnessing its internal deep feature representations as a proxy to track changes in the data distribution during deployment and, hence, to anticipate classifier performance degradation. Further, our approach harnesses the classifier's implicit dimensionality reduction of the input feature space and, hence, allows for the use of highly efficient, edge computing capable outlier detection algorithms for data drift recognition, resulting in detection rates of false negative predictions of, on average, 82 to 95 percent in our experiments. These experiments were performed on two datasets related to automated, visual quality inspection. In future work, we would like to extend experimental testing to a wider variety of data types and classifier architectures. Further, we plan to analyze the value of model supervision for other relevant tasks in automated quality inspection, that is defect localisation and segmentation.

We see our approach as a generalizable tool to assist in the manufacturing process, avoiding covert mis-predictions and reducing time and costs wasted by allowing for an early response to classifier degradation e.g. due to machine malfunctioning, sensor misreadings, environmental effects or workpiece quality related emergence of drift in the data distribution. In the words of chess grandmaster Garry Kasparov who famously lost to IBM's 'Deep Blue' computer in 1997: *"Human plus machine means finding a better way to combine better interfaces and better processes."* Kasparov, 2019